\documentclass[11pt]{article}

\usepackage[final]{acl}

\usepackage{graphicx}
\usepackage{amsmath}
\usepackage{booktabs}
\usepackage{amssymb}
\usepackage{tabularx}
\usepackage{microtype}
\usepackage[table]{xcolor}
\usepackage{multirow}
\usepackage{rotating}
\usepackage{cuted}
\usepackage{capt-of}
\usepackage{todonotes}
\usepackage{array}     
\usepackage{graphicx}
\usepackage{algorithm}
\usepackage{algpseudocode} 
\usepackage{fancyvrb}
\usepackage{xcolor} 
\newcolumntype{C}[1]{>{\centering\arraybackslash}m{#1}}
\usepackage[most]{tcolorbox}
\usepackage{subcaption}

\definecolor{promptbg}{RGB}{244,244,244}

\newenvironment{promptverbatim}
{\VerbatimEnvironment
 \begin{tcolorbox}[colback=promptbg,colframe=promptbg,boxrule=0pt,arc=2pt,
   left=4pt,right=4pt,top=3pt,bottom=3pt]
 \footnotesize\color{gray}\begin{Verbatim}}
{\end{Verbatim}\end{tcolorbox}}

\definecolor{ServOrange}{HTML}{E6550D}
\colorlet{rankFirst}{ServOrange!70}  
\colorlet{rankSecond}{ServOrange!55}
\colorlet{rankThird}{ServOrange!15}   

\usepackage[english,bidi=default]{babel} 
\babelfont{rm}{TeXGyreTermesX} 
\babelprovide[import]{hindi}
\babelfont[*devanagari]{rm}{Lohit Devanagari}
\babelprovide[import]{arabic}
\babelfont[*arabic]{rm}{Noto Sans Arabic}

\title{FineState-Bench: Benchmarking State-Conditioned Grounding for Fine-grained GUI State Setting}

\author{
\textbf{Fengxian Ji}\textsuperscript{1,2}\thanks{These authors contributed equally.},
\textbf{Jingpu Yang}\textsuperscript{2}\footnotemark[1],
\textbf{Zirui Song}\textsuperscript{1}\footnotemark[1],
\textbf{Yuanxi Wang}\textsuperscript{2}, \\
\textbf{Zhexuan Cui}\textsuperscript{2},
\textbf{Yuke Li}\textsuperscript{2},
\textbf{Qian Jiang}\textsuperscript{2},
\textbf{Xiuying Chen}\textsuperscript{1}\thanks{Corresponding author.} \\[2pt]
\textsuperscript{1}MBZUAI, United Arab Emirates \\
\textsuperscript{2}Northeastern University, China \\[3pt]
\texttt{\{fengxian.ji, zirui.song, xiuying.chen\}@mbzuai.ac.ae} \\
\texttt{\{jingpuyang290, yuanxiwang89\}@gmail.com} \\
\texttt{202219047@stu.neuq.edu.cn,202316187@stu.neu.edu.cn,llylykykk@outlook.com} \\
}

\begin{document}

\maketitle

\begin{abstract}
Despite the rapid progress of large vision-language models (LVLMs), fine-grained, state-conditioned GUI interaction remains challenging. Current evaluations offer limited coverage, imprecise target-state definitions, and an overreliance on final-task success, obscuring where and why agents fail.
To address this gap, we introduce \textbf{FineState-Bench}, a benchmark that evaluates whether an agent can correctly ground an instruction to the intended UI control and reach the exact target state.
FineState-Bench comprises 2,209 instances across desktop, web, and mobile platforms, spanning four interaction families and 23 UI component types, with each instance explicitly specifying an exact target state for fine-grained state setting.
We further propose \textit{FineState-Metrics}, a four-stage diagnostic pipeline with stage-wise success rates: Localization Success Rate (SR@Loc), Interaction Success Rate (SR@Int), Exact State Success Rate at Locate (ES-SR@Loc), and Exact State Success Rate at Interact (ES-SR@Int), 
and a plug-and-play \textit{Visual Diagnostic Assistant} (VDA) that generates a Description and a bounding-box Localization Hint to diagnose visual grounding reason via controlled w/ vs.\ w/o comparisons.
On FineState-Bench, exact goal-state success remains low: ES-SR@Int peaks at 32.8\% on Web and 22.8\% on average across platforms. With VDA localization hints, Gemini-2.5-Flash gains +14.9 ES-SR@Int points, suggesting substantial headroom from improved visual grounding, yet overall accuracy is still insufficient for reliable fine-grained state-conditioned interaction
\href{https://github.com/FengxianJi/FineState-Bench}{Github.}

\end{abstract}

\section{Introduction}
\label{sec:intro}

\begin{figure*}[t]
    \centering
    \includegraphics[width=\textwidth]{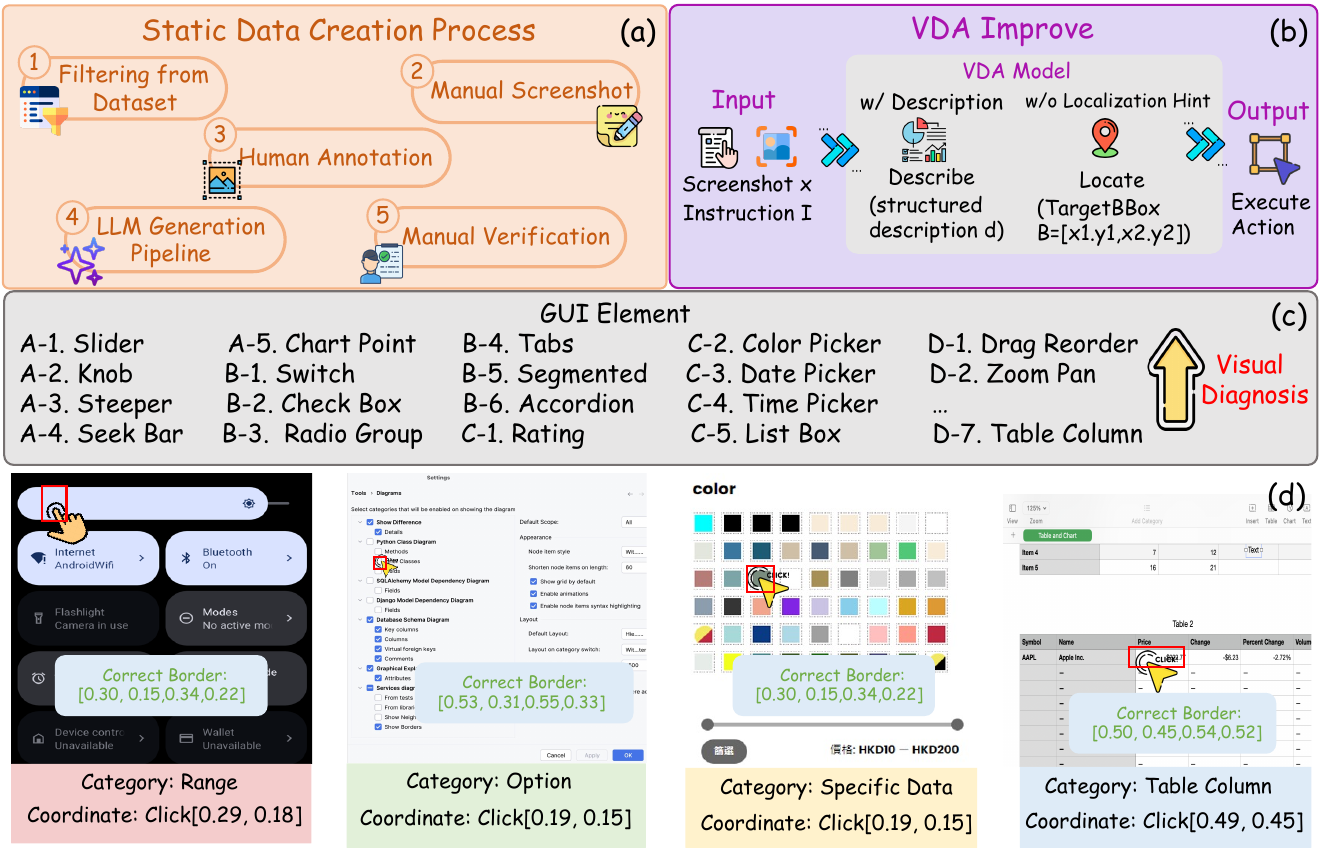}
\caption{
        \textbf{FineState-Bench overview and VDA.}
        (a) Static data creation pipeline (filtering, supplementation, annotation, instruction/state drafting, verification).
        (b) VDA-assisted evaluation that appends target-region localization hints for controlled comparisons.
        (c) Fine-grained interaction taxonomy with four families and 23 UI component types.
        (d) Example instances with precise target boxes and normalized interaction points.
    }
    \label{fig:FineState-Bench-1}
\end{figure*}

Recent advances in LVLMs have enabled a new class of GUI agents that can execute natural-language instructions on real-world software interfaces~\cite{nguyen2024improvedguigroundingiterative,wen2024autodroidllmpoweredtaskautomation}.  
By integrating visual understanding with language-conditioned decision making, such agents have shown promising capability in operating complex applications across desktop, web, and mobile environments. Representative systems such as CogAgent~\cite{hong2024cogagent} and AppAgent further demonstrate the potential of this paradigm for practical human computer interaction. To support systematic progress, prior benchmarks including AITW~\cite{gur2024aitw,ma2026medla} and ScreenSpot~\cite{you2024ferretuigroundedmobileui,qin2025uitarspioneeringautomatedgui} have provided standardized testbeds for evaluation.

Despite rapid progress in LVLM-based GUI agents \cite{zhang2025appagent,ma2026medla}, 
achieving fine-grained state-conditioned interaction remains challenging in practice. 
In many real applications, a single instruction requires setting a UI control to an exact target state using only the agent’s first predicted interaction point,
such as adjusting a slider to a precise value, selecting an exact date/time, or choosing a specific color. 
However, agents that perform strongly under coarse success criteria or standard grounding benchmarks such as ScreenSpot/SeeClick~\cite{cheng2024seeclick}, Android in the Wild~\cite{rawles2023android}, and VisualWebArena~\cite{koh-etal-2024-visualwebarena} can still fail to reliably reach precise target states.

A key limitation lies in current evaluation practices.
First, at the benchmark and task-definition level, existing evaluations for GUI and web/mobile agents predominantly focus on end-to-end task completion or click-level grounding. This design under-represents state-conditioned interaction scenarios, and the target specifications are often insufficiently precise to enable unambiguous verification of intermediate or final target states \cite{rawles2023android,deng2023mind2web,zhou2023webarena,koh-etal-2024-visualwebarena,lu2024omniparserpurevisionbased,zeng-etal-2025-bridging, cao2026taskspecificefficiencyanalysissmall}.
Second, at the evaluation-protocol and metric level, many studies primarily report aggregate outcomes such as final task success or overall accuracy. Such coarse metrics collapse the entire perception–grounding–interaction pipeline into a single score, obscuring where failures occur and preventing fine-grained diagnostic analysis and failure attribution \cite{koh-etal-2024-visualwebarena,xue2025illusionprogressassessingcurrent,emodelsyet,liang,yang2025agent}.
As a consequence, prior work may arrive at inconsistent conclusions regarding failure sources.

To address these limitations, we introduce FineState-Bench, a cross-platform benchmark of 2,209 instances for single-step, fine-grained state-conditioned GUI exact state setting with exact goal-state verification. Rather than modeling full interactive trajectories, FineState-Bench focuses on a controlled static single-step setting to isolate fine-grained state-conditioned grounding and exact state-setting ability.
Agents are evaluated under a single-step, point-based protocol, as illustrated in Figure~\ref{fig:FineState-Bench-1} and Figure~\ref{sec: ServImageModel}.
Unlike proxy-based evaluations, each instance provides exact goal-state labels and dual-region annotations (a control-extent locate box and an interactable-core box) under current/target configurations, enabling unambiguous verification of exact goal-state attainment.
Built on these annotations, we propose FineState-Metrics, a stage-wise diagnostic pipeline (SR@Loc, SR@Int, ES-SR@Loc, ES-SR@Int) that decomposes performance from component grounding to interactable-core grounding and point precision, and finally exact goal-state attainment.
To further quantify the grounding errors, we introduce a plug-and-play Visual Diagnostic Assistant (VDA) that optionally appends a Description and/or a Localization Hint; experiments show that while current agents exhibit a low success floor, VDA yields substantial gains, indicating insufficient accuracy for broad fine-grained state-conditioned interactions.

Our main contributions are:
(1) We define and study fine-grained, state-conditioned GUI state setting with explicit goal-state labels and exact verification in desktop/web/mobile platforms.
(2) We build and release FineState-Bench featuring dual-region annotations under current/target configurations for component localization and interactable-core precision, enabling fine-grained, reproducible evaluation.
(3) We introduce FineState-Metrics and VDA to decompose failures and perform controlled input-augmentation analysis, and we benchmark 8 representative agents to quantify how much performance is limited by interactable-core localization and point precision.

\section{Related Work}
\label{sec: Related Work}

\paragraph{GUI Agents.}
GUI agents have advanced rapidly with large vision-language and multimodal models \cite{zhang2024large,nguyen2024guiagents}.
Early studies often relied on general-purpose models for GUI operation \cite{yang2023setofmark,apt,zheng2024gpt4v}, while later work increasingly builds GUI-specialized agents and UI grounding models \cite{hong2024cogagent,screenagent,lin2024showui,OS-Atlas,Li2025FerretUI2,gou2024uground,chen2024guicourse,qin2023rethinking}.
These systems span mobile and desktop platforms \cite{wang2024mobile,Nong2024MobileFlow,jiang2025appagentxevolvingguiagents,liu2025pc,Jedi-7B-1080p,zhang2025ufo,yang2025frequencypointgameenvironment,yang2026frequency}, yet reliably achieving state-conditioned exact state setting remains challenging.
In practice, small execution errors can accumulate into user-visible failures, especially for precise controls such as sliders, pickers, and professional UI widgets, motivating evaluations that stress fine-grained state manipulation rather than only task completion.

\paragraph{GUI Agent Evaluation.}
Existing evaluations include interactive end-to-end benchmarks \cite{zhou2023webarena,koh-etal-2024-visualwebarena,rawles2024androidworld,xie2024osworld,zhang2024a-star,chen2025spabench,zhang2024llamatouch}
and offline grounding benchmarks on screenshots or professional software \cite{deng2023mind2web,zhao2024screenspotpro,qian2024visual,dardouri2024visual}.
While these benchmarks improve realism or grounding assessment, they often under-cover state-conditioned interaction scenarios and lack precise target-state specifications for unambiguous verification.
Moreover, many protocols emphasize aggregate success rates, limiting deeper analysis \cite{zheng2024guirobust,zhang2024worldgui}.
Recent benchmarks probe robustness or distribution shifts, but they typically do not isolate failures from mis-perception, mis-localization, or incorrect state outcomes when exact goal-state attainment is required.

\paragraph{Diagnosis and Failure Attribution.}
Recent work highlights that coarse metrics can obscure bottlenecks and yield divergent conclusions about failure sources \cite{Shlomov2024GroundingPlanning}.
Related efforts improve grounding or data coverage \cite{Chen2024EDGE} or analyze reliability issues \cite{liu2024agentsmith,liang2025vision}, but controlled diagnosis for disentangling visual grounding from non-visual interaction and state-control factors remains limited.

\section{FineState-Bench}
\label{sec:FineState-Bench}
\subsection{Problem Definition}
\label{sec:problem_definition}
FineState-Bench targets fine-grained, state-conditioned GUI state setting with exact goal states.
Given a screenshot $x$ and two instructions $(I^{0}, I^{1})$, an agent predicts two points $(p^{0}, p^{1})$: $p^{0}$ indicates the target control's location in the current UI, and $p^{1}$ indicates the operation location required to reach the intended fine-grained goal state.
Specifically, $p^{0}$ corresponds to the current instruction $I^{0}$ for locating the target control, whereas $p^{1}$ corresponds to the target instruction $I^{1}$ for specifying the operation location toward the intended goal state.
Each instance is defined as $\tau = (x, I^{0}, I^{1}, c^*, s_{\mathrm{goal}}, B_{\mathrm{loc}}^{0}, B_{\mathrm{int}}^{0}, B_{\mathrm{loc}}^{1}, B_{\mathrm{int}}^{1})$, where $c^*$ is the target control and $s_{\mathrm{goal}}$ is the desired goal state.
All boxes are axis-aligned rectangles in normalized screen coordinates, each parameterized by $(x_{\min}, y_{\min}, x_{\max}, y_{\max}) \in [0,1]^4$; for trajectory-based actions, we define $p^{t}$ ($t\in\{0,1\}$) as the final release (action-commit) point, since it determines the resulting state in typical GUI systems.

$B_{\mathrm{loc}}^{0}$ denotes the locate box covering the current visible extent of $c^*$, while $B_{\mathrm{int}}^{0}$ denotes the box of the operation-relevant element/region in the current configuration for reaching $s_{\mathrm{goal}}$.
To account for controls whose position or size may change during the intended operation, we additionally annotate target-configuration boxes:
$B_{\mathrm{loc}}^{1}$ denotes the locate box of $c^*$ in the target configuration, and $B_{\mathrm{int}}^{1}$ denotes the corresponding box of the operation-relevant element/region in the target configuration for reaching $s_{\mathrm{goal}}$.
We write $p \in B$ when a predicted point $p$ falls inside box $B$; we compare $p^{0}$ against $(B_{\mathrm{loc}}^{0}, B_{\mathrm{loc}}^{1})$ and $p^{1}$ against $(B_{\mathrm{int}}^{0}, B_{\mathrm{int}}^{1})$.
This enables evaluation under both the current ($\cdot^{0}$) and target ($\cdot^{1}$) configurations, capturing potential layout changes during the intended operation.

Each control $c$ has a precise, quantifiable state $s(c)$, such as a slider value, toggle status, selected option. 
This two-point design decouples locating the target control in the current UI ($p^{0}$) from specifying the goal-directed operation location ($p^{1}$), reducing interference from multi-step interactions and enabling fine-grained diagnosis of localization versus operation.

\subsection{Dataset Construction and Annotation}
\paragraph{Benchmark Composition.}
FineState-Bench contains 2,209 high-quality static GUI state-setting instances collected from three platforms: Desktop (810), Web (701), and Mobile (698), with data sourcing and quality control detailed in Appendix~\ref{app:data_collection}.
Each instance includes: (i) a screenshot $x$, (ii) two instructions $(I^{0}, I^{1})$ for current localization and target operation specification, (iii) an exact goal-state label $s_{\mathrm{goal}}$ with fine-grained state annotations for the target control, and (iv) four geometric annotations for the target control in the current and target configurations $(B_{\mathrm{loc}}^{0}, B_{\mathrm{int}}^{0}, B_{\mathrm{loc}}^{1}, B_{\mathrm{int}}^{1})$.
We further verify the reliability of the fine-grained state and geometric annotations through a double-annotation study; details are provided in Appendix~\ref{app:Annotation Reliability}.

\paragraph{Interaction Taxonomy.}
To ensure systematic coverage, we group instances into four interaction families:
(1) Numerical and Range Adjustment;
(2) State Toggling and Option Selection;
(3) Specific Data-Type Selection;
and (4) Content Organization and View Manipulation.

\begin{figure}[t]
    \centering
    \begin{minipage}[t]{0.22\textwidth}
        \centering
        \includegraphics[width=\textwidth]{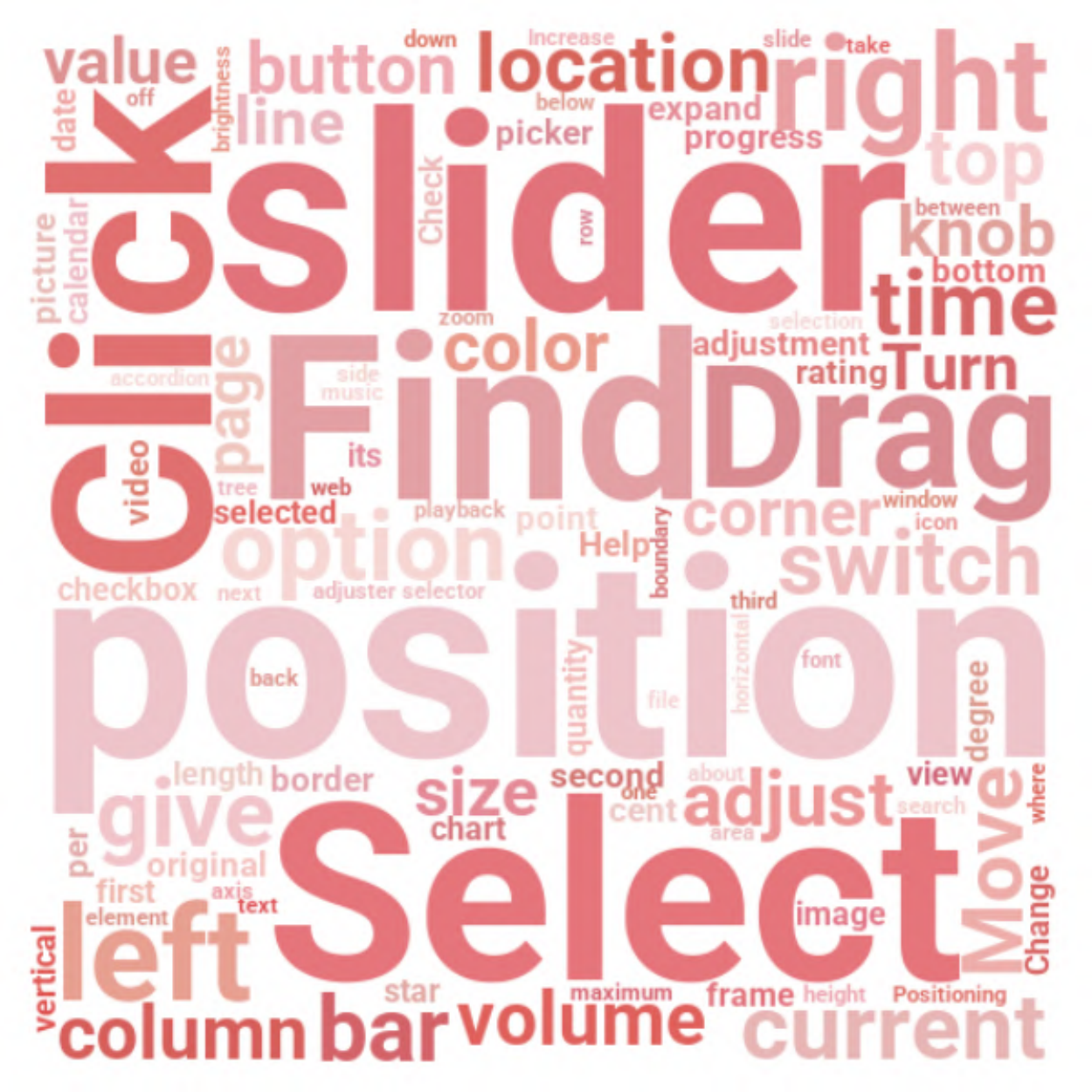}
        \caption*{(a) Word Cloud}
    \end{minipage}
    \hfill
    \begin{minipage}[t]{0.24\textwidth}
        \centering
        \includegraphics[width=\textwidth]{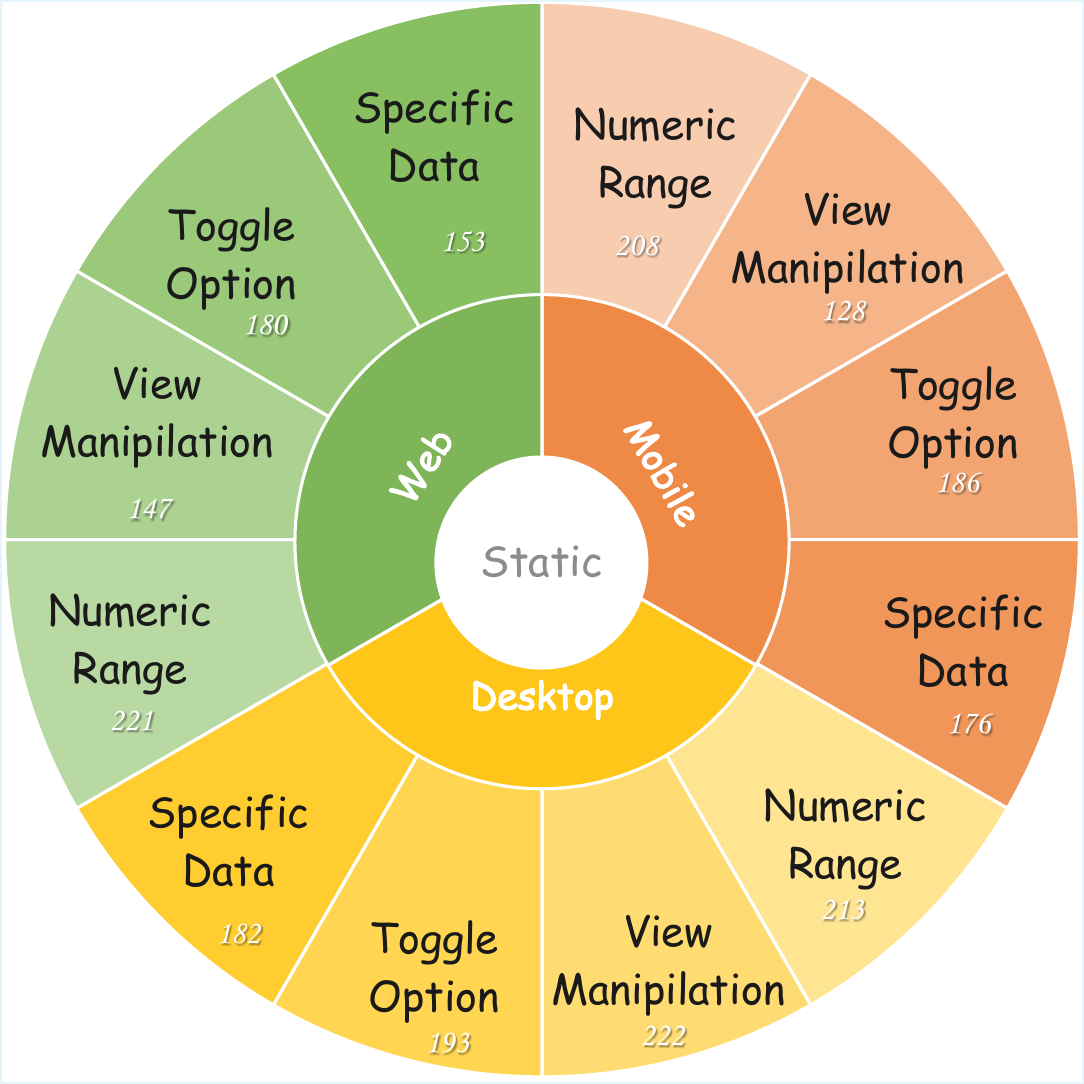}
        \caption*{(b) Task Type Distribution}
    \end{minipage}
    \caption{Visualization of instruction keywords and task type composition in FineState-Bench.}
    \label{fig:wordcloud_pie}
\end{figure}

\paragraph{Instruction Statistics.}
Figure~\ref{fig:wordcloud_pie}(a) summarizes instruction keywords, highlighting common interaction verbs such as select, click, drag, and adjust, alongside state-related entities including volume, brightness, time, and color.
Figure~\ref{fig:wordcloud_pie}(b) reports the task-type distribution across the benchmark.

\begin{table*}[!t]
\centering
\scriptsize
\renewcommand{\arraystretch}{0.95}
\resizebox{\textwidth}{!}{%
\begin{tabular}{c ccc ccc c} 
\toprule
\multirow{2}{*}{\textbf{Benchmark}}
& \multicolumn{3}{c}{\textbf{Platform}}
& \multicolumn{3}{c}{\textbf{State-Control Evaluation Properties}} 
& \multirow{2}{*}{\textbf{Number}} \\
\cmidrule(lr){2-4} \cmidrule(lr){5-7} 
& Desktop & Mobile & Website
& Target State Labels & $B_{\text{loc}}/B_{\text{int}}$ & Stage-wise Diag
& \\
\midrule
ScreenSpot~\cite{cheng2024seeclick}
& Yes & Yes & Yes
& No  & No & No
& 1272 \\
ScreenSpot-v2~\cite{OS-Atlas}
& Yes & Yes & Yes
& No  & No & No
& 1272 \\
ScreenSpot-Pro~\cite{zhao2024screenspotpro}
& Yes & --  & --
& No  & No & No
& 1581 \\
WebClick~\cite{andreux2025surfer}
& --  & --  & Yes
& No  & No & No
& 1639 \\
VisualWebBench~\cite{liu2024visualwebbench}
& --  & --  & Yes
& No  & No & No
& 1536 \\
UI-Vision~\cite{nayak2025ui}
& Yes & --  & --
& No  & No & No
& 1464 \\
OSWorld-G~\cite{OSWORLD-G}
& Yes & --  & --
& No  & No & Yes
& 564 \\
\midrule
FineState-Bench
& Yes & Yes & Yes
& Yes & Yes & Yes
& 2209 \\
\bottomrule
\end{tabular}%
}
\caption{
Comparison of FineState-Bench with representative static GUI benchmarks in terms of evaluation metrics and diagnostic capabilities. 
We summarize whether each benchmark supports Target State Labels verification, dual-region geometric supervision ($B_{\text{loc}}/B_{\text{int}}$), and stage-wise diagnostic metrics that disentangle localization accuracy, point-level precision, and exact state attainment.
}
\label{tab:dataset_comparison}
\end{table*}

\subsection{FineState-Metrics}

While overall task success provides a coarse measure of agent capability, it offers limited insight into where fine-grained state control fails along the perception-to-interaction pipeline.
FineState-Bench is designed not only to score overall success, but also to attribute failures to specific capability factors in fine-grained state control. 
As defined in \S\ref{sec:problem_definition}, we evaluate each instance using the two predicted points $(p^{0}, p^{1})$ and the resulting (recorded/annotated) target state $s_1(c^*)$.

We introduce four diagnostic Success Rates aligned with the four pipeline stages of Perception, Localization, Interaction, and State Correctness.
Perception Success Rate measures instruction understanding and goal identification by checking whether the agent correctly identifies the target control and intended goal state from the inputs $(I^{0}, I^{1})$, independent of any point prediction.
Localization Success Rate (SR@Loc) measures coarse component grounding (control  by checking whether the predicted control-location point $p^{0}$ falls inside the locate box of the target control in the current configuration, i.e., $p^{0}\in B_{\mathrm{loc}}^{0}$, 
while Interaction Success Rate (SR@Int) measures fine-grained interactable-core grounding and point precision by checking whether the predicted operation point $p^{1}$ falls inside the operation-relevant (state-changing) region in the current configuration, i.e., $p^{1}\in B_{\mathrm{int}}^{0}$.
Exact State Success Rate (ES-SR) measures exact goal-state attainment by checking whether the resulting (recorded/annotated) state of the target control reaches the exact goal state, $s_1(c^*) = s_{\mathrm{goal}}$.
Accordingly, Exact State Success Rate at Locate (ES-SR@Loc) and Exact State Success Rate at Interact (ES-SR@Int) measure exact goal-state success conditioned on $p^{0}\in B_{\mathrm{loc}}^{1}$ and $p^{1}\in B_{\mathrm{int}}^{1}$, respectively, i.e., under the target-configuration locate/interact boxes (see Appendix~\ref{app:metric_interpretation} for interpretation and failure-attribution rules):
\begin{equation}
\scalebox{0.67}{$
\left\{
\begin{aligned}
\mathrm{SR@Loc} &= \frac{1}{N}\sum_{i=1}^{N}\mathbb{I}\!\left[p^{(i)}_{0}\in B^{(i)}_{\mathrm{loc}}{}^{0}\right],\\
\mathrm{SR@Int} &= \frac{1}{N}\sum_{i=1}^{N}\mathbb{I}\!\left[p^{(i)}_{1}\in B^{(i)}_{\mathrm{int}}{}^{0}\right],\\
\mathrm{ES\text{-}SR@Loc} &= \frac{1}{N}\sum_{i=1}^{N}\mathbb{I}\!\left[p^{(i)}_{0}\in B^{(i)}_{\mathrm{loc}}{}^{1}
\ \wedge\ s^{(i)}_{1}\!\left(c^{*(i)}\right)=s^{(i)}_{\mathrm{goal}}\right],\\
\mathrm{ES\text{-}SR@Int} &= \frac{1}{N}\sum_{i=1}^{N}\mathbb{I}\!\left[p^{(i)}_{1}\in B^{(i)}_{\mathrm{int}}{}^{1}
\ \wedge\ s^{(i)}_{1}\!\left(c^{*(i)}\right)=s^{(i)}_{\mathrm{goal}}\right].
\end{aligned}
\right.
$}
\label{eq:finestate-metrics}
\end{equation}
Here $N$ is the number of instances, $(i)$ indexes an instance, $\mathbb{I}(\cdot)$ is the indicator function, and $\wedge$ denotes conjunction.

For example, in Fig.\ref{fig:FineState-Bench} (D5), SR@Loc is counted as success if $p^{0}$ lands within the TreeView's current visible extent ($B_{\mathrm{loc}}^{0}$), whereas SR@Int requires $p^{1}$ to hit the current operation-relevant region ($B_{\mathrm{int}}^{0}$) that can trigger the intended state change.
If the agent hits $B_{\mathrm{int}}^{1}$ but the resulting state still differs from $s_{\mathrm{goal}}$, ES-SR@Int remains unsuccessful, indicating a state-setting error beyond location/operation prediction.

\begin{figure}[t]
    \centering
    \includegraphics[width=\linewidth]{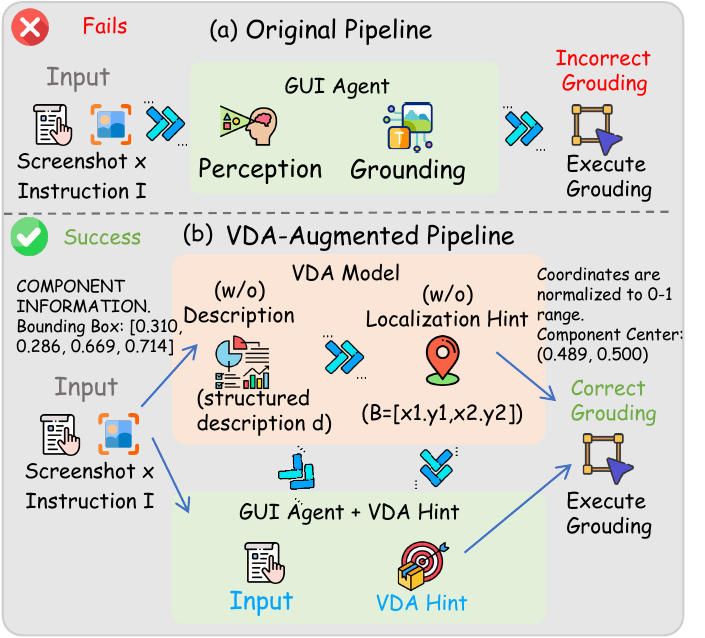}
    \caption{Baseline and VDA-augmented pipelines for the target instruction $I^{1}$. VDA (GPT-4o) first produces a structured Description of the target UI element and then a Localization Hint $\hat{B}^{1}$, which is appended to the agent input when predicting $p^{1}$. The performance gap in ES-SR@Int quantifies the visual grounding bottleneck for goal-directed operation.}
    \label{fig:vda_pipeline}
\end{figure}

\subsection{Visual Diagnostic Assistant}

\paragraph{Diagnostic Use of VDA.}
To conduct an in-depth study of the factors influencing errors in fine-grained operations, we introduce the Visual Diagnostic Assistant (VDA) as a diagnostic tool for controlled comparisons.
We use w/ and w/o to denote with and without VDA.
VDA evaluates the same agent with and without an explicit Localization Hint under the target instruction $I^{1}$, isolating whether failures arise from inaccurate grounding of the goal-directed operation region or from visual perception factors.
As shown in Figure~\ref{fig:vda_pipeline}, VDA produces a structured Description and a Localization Hint (a target-region bounding box) for $I^{1}$; either the Description, the Localization Hint, or both are appended to the agent input.

Concretely, given $(x, I^{1})$, VDA predicts $\hat{B}^{1}\in[0,1]^4$ as the Localization Hint: $\hat{B}^{1} = \mathcal{L}(x, I^{1})$, where $\mathcal{L}$ is instantiated with GPT-4o.
Using ES-SR@Int under $I^{1}$, we quantify the visual grounding bottleneck via the w/ vs.\ w/o VDA gap:
\begin{equation}
\resizebox{0.895\linewidth}{!}{$
\Delta_{\mathrm{vis}} = \mathrm{ES\text{-}SR@Int}(\text{w/ VDA})
- \mathrm{ES\text{-}SR@Int}(\text{w/o VDA}).
$}
\end{equation}
$\Delta_{\mathrm{vis}}$ estimates the performance recoverable from improved visual grounding under this Localization Hint interface.

\paragraph{VDA Design.}
VDA follows a two-step describe-then-localize procedure under $I^{1}$ to produce a high-fidelity Localization Hint.
First, given $(x, I^{1})$, VDA generates a Description of the target UI element, including its functional role or state, discriminative visual cues, and spatial relations to anchors.
This Description is used for disambiguation and is optionally provided to the agent.
Then, conditioned on the screenshot, the instruction $I^{1}$, and the generated Description, VDA predicts $\hat{B}^{1}$ in normalized coordinates as the Localization Hint.

\paragraph{Plug-and-Play Integration of VDA.}
For each instance, when predicting $p^{1}$ under $I^{1}$, the evaluated agent receives its standard inputs optionally augmented with the VDA-predicted Localization Hint $\hat{B}^{1}$.
Importantly, the intervention can be the Description, the Localization Hint, or both.
This design enables controlled ablations that isolate the effect of the Localization Hint from textual disambiguation.

\section{Benchmark Characteristics and Analysis}
\label{sec: ServImageModel}

\begin{figure*}[!t]
    \centering
    \includegraphics[width=0.98\textwidth]{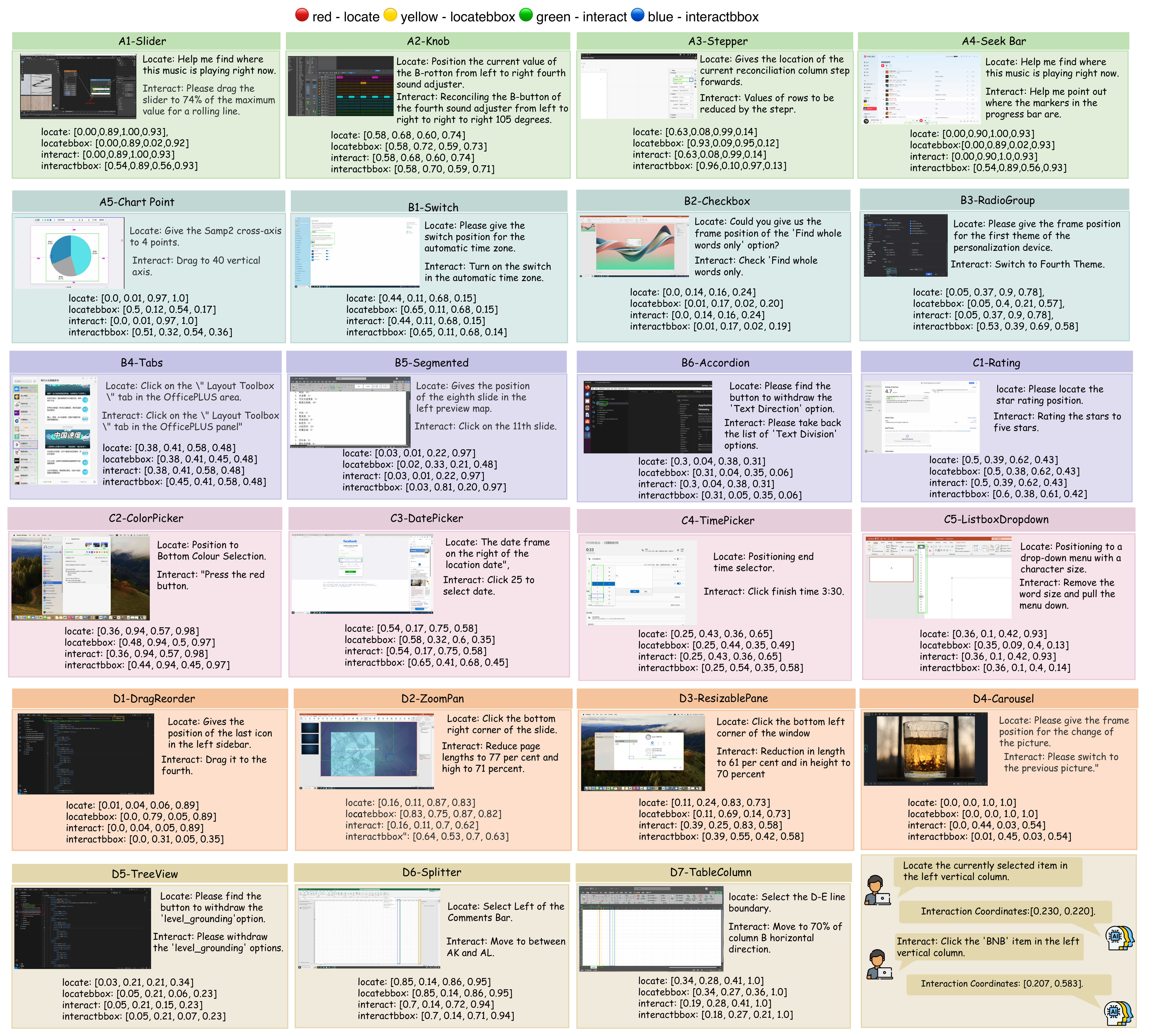}
    \caption{Representative instances from FineState-Bench, covering all 23 UI component types. For each control, we annotate four normalized bounding boxes.
    }
    \label{fig:FineState-Bench}
\end{figure*}

Table~\ref{tab:dataset_comparison} shows that representative offline/static GUI benchmarks mainly test whether an agent can identify and click the intended element, or respond to higher-level prompts, without requiring verifiable post-interaction state changes. 
As a result, models may score well even if the interaction is not precise enough to reach an exact target state.
Moreover, they often lack a clear distinction between a control’s visible extent and the operation-relevant region that changes its state, making failures hard to interpret whether the model mis-grounded the control, clicked an ineffective region, or failed to set the correct state.

FineState-Bench addresses this gap by evaluating whether an agent can set a target control to an exact goal state with a single-step, single-point interaction.
Each instance provides a goal state and dual bounding-box supervision, as shown in Figure~\ref{fig:FineState-Bench}: the locate box covers the control’s extent, while the interact box marks the state-changing core region.
This design enables clear attribution in a single interaction: missing the locate box indicates grounding failure; hitting locate but missing interact suggests insufficient point precision; and hitting interact but not reaching the goal state indicates a state-setting error beyond localization.
\begin{table*}[!tp]
\centering
\scriptsize
\setlength{\tabcolsep}{3pt}
\renewcommand{\arraystretch}{0.9}
\resizebox{\textwidth}{!}{%
\begin{tabular}{lcccc} %
\toprule
\textbf{Model Name} & \textbf{Mobile} & \textbf{Web} & \textbf{Desktop} & \textbf{AVG} \\
\midrule
\multicolumn{5}{c}{\textit{\textbf{Closed-source Models}}} \\
\midrule
GPT-4o & 31.0/6.2/9.1/6.2 & 22.8/4.5/7.0/4.5 & 20.6/2.2/4.4/2.2 & 24.8/4.3/6.8/4.3 \\
Claude-3.5-Sonnet & 31.7/11.5/13.7/11.5 & 15.2/1.9/4.6/1.9 & 22.0/3.4/5.4/3.4 & 23.0/5.6/7.9/5.6 \\
Gemini-2.5-Flash & 49.4/17.6/21.1/17.6 & 47.0/11.8/16.8/11.8 & 12.7/0.7/3.8/0.7 & 36.4/10.0/13.9/10.0 \\
\midrule
\multicolumn{5}{c}{\textit{\textbf{Open-source Models}}} \\
\midrule
OS-Atlas-7B \cite{OS-Atlas} & 47.5/12.8/18.7/12.8 & 33.2/7.5/9.2/7.5 & 45.3/9.8/15.2/9.8 & 42.0/10.0/14.4/10.0 \\
CogAgent-9B \cite{hong2024cogagent} & 17.7/1.8/2.5/1.8 & 24.1/6.4/13.7/6.4 & 29.4/2.3/3.5/1.2 & 23.7/3.5/6.6/3.1 \\
UGround-7B \cite{gou2024uground} & 50.7/19.6/22.4/19.6 & 62.0/32.8/62.0/32.8 & 46.3/16.0/25.4/16.0 & 53.0/22.8/36.6/22.8 \\
Jedi-7B-1080p \cite{Jedi-7B-1080p} & 13.3/1.6/3.1/1.5 & 12.7/8.3/12.7/8.3 & 12.2/0.8/1.5/0.8 & 12.7/3.6/5.8/3.5 \\
ShowUI-2B \cite{lin2024showui} & 20.3/5.2/6.7/5.2 & 26.7/5.3/26.7/5.3 & 30.3/3.2/9.1/3.2 & 25.8/4.6/14.2/4.6 \\
\bottomrule
\end{tabular}%
}
\caption{Baseline evaluation on FineState-Static. Under the single-point protocol, we report four diagnostic metrics:
SR@Loc ($p^{0}\!\in\!B_{\mathrm{loc}}^{0}$) / SR@Int ($p^{1}\!\in\!B_{\mathrm{int}}^{0}$) / ES-SR@Loc ($p^{0}\!\in\!B_{\mathrm{loc}}^{1}$) / ES-SR@Int ($p^{1}\!\in\!B_{\mathrm{int}}^{1}$) (\%).}
\label{tab:static_results_en_corrected}
\end{table*}

\section{Experiments and Analysis}
\label{sec: Experiments and Results}

\subsection{Baselines}
We benchmark 8 representative GUI agents on FineState-Static. 
For a balanced comparison, we include 3 closed-source LVLMs (GPT-4o~\cite{gpt4o}, Claude-3.5-Sonnet~\cite{claude_3_5_sonnet}, Gemini-2.5-Flash~\cite{gemini_2_5_flash}) as strong general-purpose multimodal baselines, and 5 open-source GUI agents (OS-Atlas-7B~\cite{OS-Atlas}, CogAgent-9B~\cite{hong2024cogagent}, UGround-7B~\cite{gou2024uground}, Jedi-7B-1080p~\cite{Jedi-7B-1080p}, ShowUI-2B~\cite{lin2024showui}) that emphasize GUI grounding and action prediction. 

We evaluate all agents under the single-point protocol, using ES-SR@Int as the primary metric for exact goal-state attainment, and leveraging the stage-wise success rates (SR@Loc, SR@Int, ES-SR@Loc) for diagnosis and bottleneck attribution.

\subsection{Main Result}
Table~\ref{tab:static_results_en_corrected} shows consistently low success for exact state setting, with ES-SR@Int as the primary metric (see Appendix~\ref{app:metric_interpretation} for metric interpretation and rule-based failure attribution).
Even the strongest model, UGround-7B, reaches only 32.8\% ES-SR@Int on Web and 22.8\% on average.
Performance can also collapse on specific platforms despite reasonable localization (e.g., Gemini-2.5-Flash: 17.6\% on Mobile vs.\ 0.7\% on Desktop), indicating that robust fine-grained state setting remains difficult even on static screenshots.

\begin{figure}[t]
    \centering
    \includegraphics[width=\linewidth]{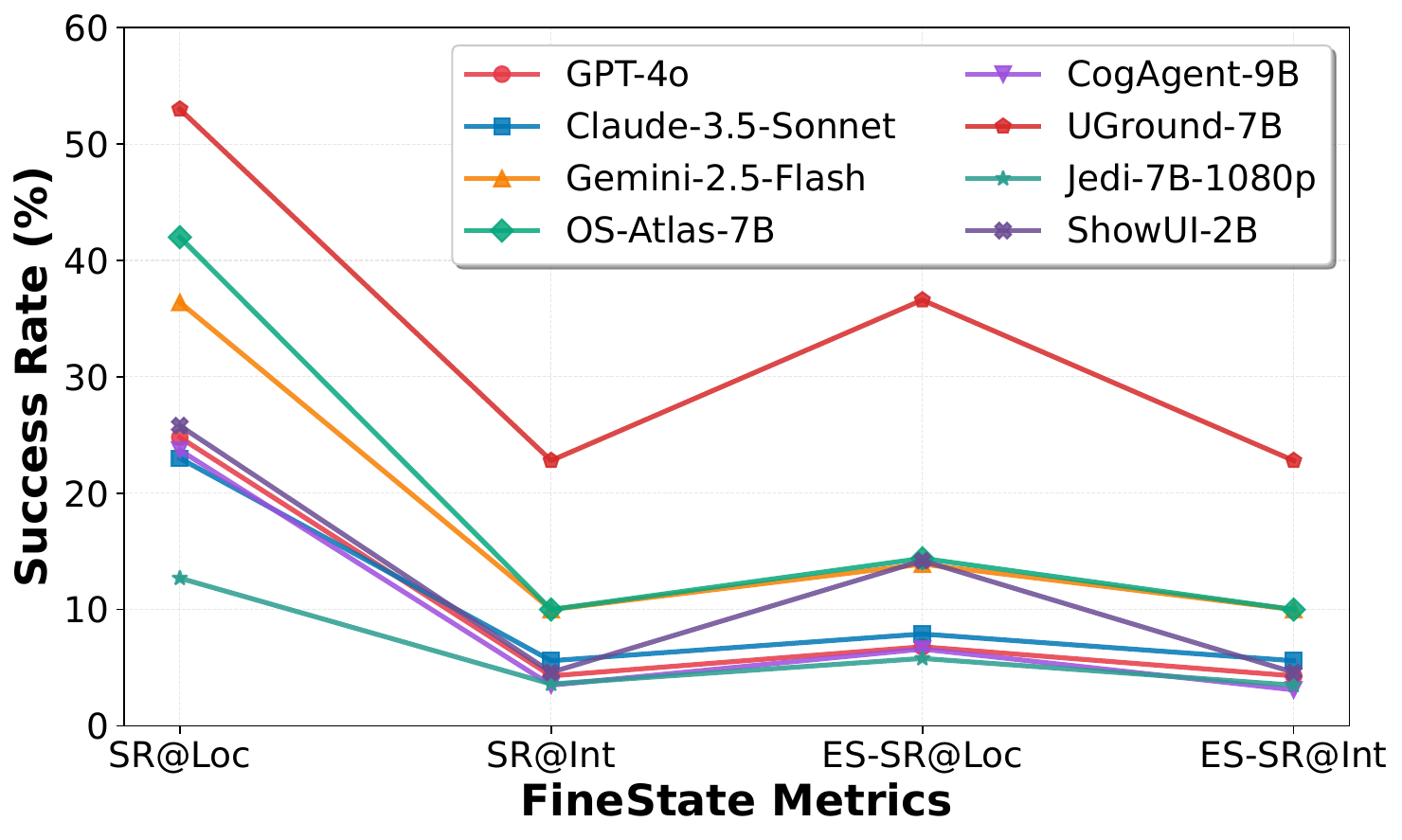}
    \caption{Performance degradation on FineState-Static.}
    \label{fig:stage_degradation}
\end{figure}

FineState-Metrics attributes most errors to the transition from coarse component grounding to interactable-core grounding.
Across models, SR@Loc is substantially higher than SR@Int, suggesting that agents often localize the correct control ($p^{0}!\in!B_{\mathrm{loc}}^{0}$) but miss the state-changing core when executing ($p^{1}!\notin!B_{\mathrm{int}}^{0}$).
Moreover, SR@Int is typically close to ES-SR@Int, implying that once the predicted point hits the interactable core, the exact goal state is usually achieved.
This stage-wise degradation is visualized in Fig.~\ref{fig:stage_degradation}, highlighting interactable-core grounding as the dominant bottleneck for broad fine-grained, state-conditioned interaction.

\subsection{Component-Level Diagnosis of Interactable-Core Grounding}

\begin{figure*}[t]
    \centering
    \includegraphics[width=\textwidth]{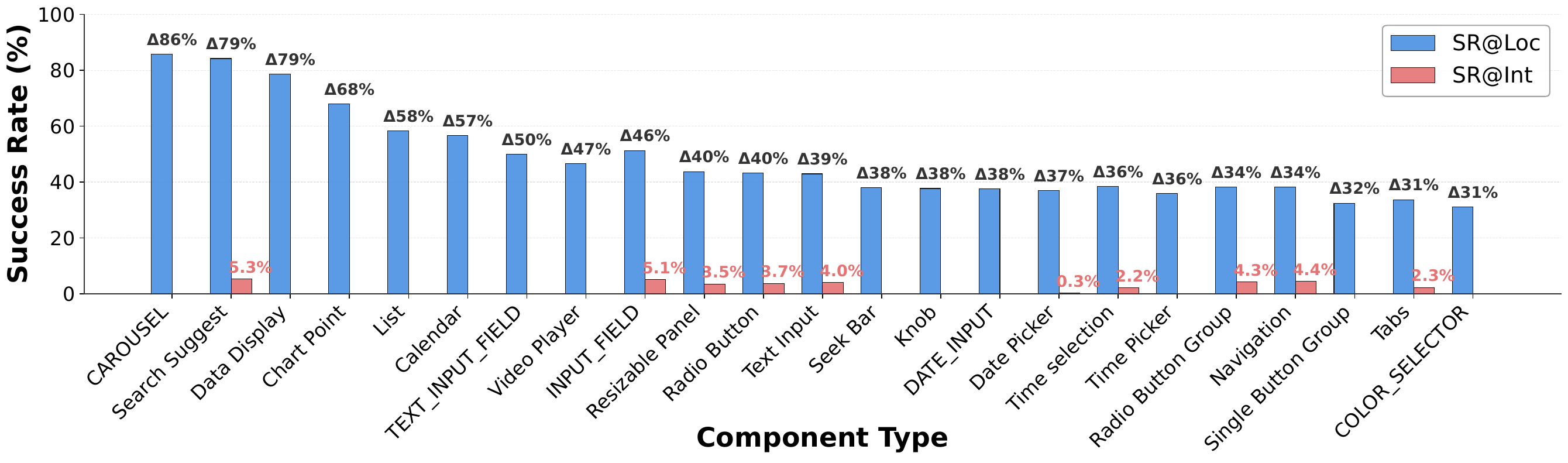}
    \caption{
   Component-level localization vs. interactable-core grounding on FineState-Static.
    }
\label{fig:component_loc_intloc}
\end{figure*}

As shown in Figure~\ref{fig:component_loc_intloc}, we break down SR@Loc and SR@Int by UI component to diagnose which interaction types contribute most to the drop from coarse localization to interactable-core grounding.
We find that precision-sensitive, continuous controls exhibit the largest gaps: on sliders, models often achieve reasonable SR@Loc but much lower SR@Int, and on seek bars SR@Int can even collapse despite strong localization.
In contrast, discrete controls are more tractable, with steppers showing a much smaller gap between SR@Loc and SR@Int.
Overall, these component-wise results suggest that the dominant bottleneck in broad fine-grained interactions is the transition from coarse component localization ($p^{0}!\in!B_{\mathrm{loc}}^{0}$) to interactable-core grounding ($p^{1}!\in!B_{\mathrm{int}}^{0}$); see Table~\ref{tab:component_breakdown} for full numbers.

\subsection{Diagnosis of Performance Bottlenecks}

To attribute failures, we run a controlled comparison with VDA, which injects full VDA hints consisting of a Description and a Localization Hint, while keeping the evaluated agent unchanged (same parameters/decoding and the same point-based interaction interface).
To attribute failures, we run a controlled comparison with VDA, which injects full VDA hint, consisting of the Description and the Localization Hint, while keeping the evaluated agent unchanged (same parameters/decoding and the same point-based interaction interface).

As shown in Table~\ref{tab:VDA_overall_en}, VDA-generated localization hints substantially improve ES-SR@Int, with the largest gain observed on Gemini-2.5-Flash (+14.9\% on average).
These recoverable gains indicate that a major portion of failures stems primarily from inaccurate grounding/localization of the goal-directed operation region, with visual perception factors playing a secondary role, while overall accuracy still remains insufficient for broad fine-grained interactions.

\begin{table}[h]
\centering
\scriptsize
\setlength{\tabcolsep}{2.5pt}
\renewcommand{\arraystretch}{0.8}
\resizebox{\columnwidth}{!}{%
\begin{tabular}{cccc}
\toprule
\textbf{Model} & \textbf{Mobile} & \textbf{Web} & \textbf{Desktop}  \\
\midrule
Gemini-2.5-Flash & 17.6 & 11.8 & 0.7  \\
ShowUI-2B     & 5.2 & 5.3 & 3.2  \\
OS-Atlas-7B    & 12.8 & 7.5 & 9.8 \\
\midrule
Gemini-2.5-Flash(VDA-Gemini-2.5-Flash) & 29.8 & 27.2 & 15.4  \\
ShowUI-2B(VDA-Gemini-2.5-Flash)     & 5.8 & 12.3 & 7.5  \\
OS-Atlas-7B(VDA-Gemini-2.5-Flash)    & 18.9 & 18.2 & 19.1 \\
\midrule
Gemini-2.5-Flash(VDA-GPT4o) & 29.8 & 26.6 & 20.9  \\
ShowUI-2B(VDA-GPT4o)     & 7.3 & 7.2 & 9.5  \\
OS-Atlas-7B(VDA-GPT4o)    & 15.9 & 14.2 & 13.1 \\

\bottomrule
\end{tabular}%
}
\caption{Overall impact of VDA on ES-SR@Int (\%). We report baseline and VDA-augmented performance under identical agent settings; improvements reflect error recoverable by better visual grounding.}
\label{tab:VDA_overall_en}
\end{table}

\subsection{Ablation Study of VDA}

We conduct an ablation study on Gemini-2.5-Flash in Table~\ref{tab:VDA_ablation_en} to each VDA component / input cue / hint type under the single-point protocol.
Using w/ Description but w/o localization hint yields no measurable improvement over w/o VDA, with ES-SR@Int remaining nearly unchanged across platforms.
By contrast, using w/ Localization Hint produces a pronounced increase in ES-SR@Int across all platforms, indicating that localization quality is a major driver of the overall gain.
The the full variant (w/ Description and w/ Localization Hint) (w/ Description and w/ Localization Hint) performs best, consistent with the description providing contextual disambiguation that improves the reliability of subsequent localization.

\begin{table}[h]
\centering
\scriptsize
\setlength{\tabcolsep}{2.5pt}
\renewcommand{\arraystretch}{0.8}
\resizebox{\columnwidth}{!}{%
\begin{tabular}{lccc}
\toprule
\textbf{VDA Configuration} & \textbf{Mobile} & \textbf{Web} & \textbf{Desktop}\\
\midrule
w/o VDA & 17.6 & 11.8 & 0.7 \\
w/ Description, w/o Localization Hint & 17.9 & 11.9 & 0.8 \\
w/o Description, w/ Localization Hint & 26.1 & 24.7 & 13.1 \\
w/ Description, w/ Localization Hint & 29.8 & 27.2 & 15.4 \\
\bottomrule
\end{tabular}%
}
\caption{Ablation study of VDA-Flash on Gemini-2.5-Flash. We measure ES-SR@Int (\%) under the single-point protocol.}
\label{tab:VDA_ablation_en}
\end{table}

We provide qualitative failure cases and w/ VDA comparisons in \autoref{app:Failure Cases} of the Supplementary materials to illustrate typical error modes, such as missing the interactable core due to imprecise point placement.
\section{Conclusion}
\label{sec:Conclusion}
We present FineState-Bench, an open-source benchmark and diagnostic framework for fine-grained, state-conditioned GUI state setting with exact goal-state verification across desktop, web, and mobile platforms. 
Our study highlights a key gap in current GUI agent evaluation: existing benchmarks rarely support exact goal-state verification and controlled, stage-wise diagnosis, making it difficult to attribute failures in fine-grained state-conditioned interactions.
We hope FineState-Bench, together with FineState-Metrics and VDA-based analysis, will enable precise evaluation and accelerate progress toward reliable, state-aware GUI agents, enabling attribution analysis.

\section*{Limitation}
Our study has several limitations.
First, FineState-Bench is designed to isolate fine-grained, state-conditioned state setting under static screenshots and a single-point interaction setting. This controlled setup prioritizes precise state verification and diagnostic clarity, rather than modeling long-horizon reasoning or multi-step corrective behaviors, which are complementary directions explored by existing interactive benchmarks.
Second, VDA is designed as a diagnostic tool rather than a deployable component, and its localization hints rely on high-quality visual cues in the screenshot; extending this analysis to fully end-to-end or real-time interactive settings remains an open challenge.
Third, while the benchmark covers a broad range of platforms and UI components, it does not exhaustively represent all application domains or accessibility-driven interface variations, which may exhibit different grounding and state-control characteristics.

\section*{Acknowledgements}
We thank the anonymous reviewers and the area chair for their constructive comments. We also thank our mentors and colleagues from MBZUAI for their support and help.

{
    \small
    \bibliography{main}
}

\appendix

\section{Data Collection and Quality Control}
\label{app:data_collection}

\paragraph{Benchmark Composition.}
FineState-Bench (FineState-Static) contains 2,209  instances across Desktop (810), Web (701), and Mobile (698).
Each instance includes a screenshot, an instruction that specifies an exact target state, fine-grained state labels, and geometric annotations (dual bounding boxes).
All coordinates are normalized to $[0,1]$.

\paragraph{Construction Pipeline.}
We curate the benchmark through LVLM-based pre-filtering and manual verification, following a five-step pipeline:
(1) Filtering from dataset: use an LVLM to pre-filter candidates with non-trivial state changes from OS-Atlas;
(2) Manual screenshot supplementation: add missing but representative interaction patterns to ensure coverage of all component types;
(3) Human annotation: annotate dual bounding boxes and state labels;
(4) LLM-assisted drafting: draft candidate instruction--state pairs, then refine them to enforce exact target states;
(5) Manual verification: validate instruction state consistency and bounding box quality, and remove ambiguous or noisy cases.

\paragraph{Interaction Taxonomy: 23 Component Types.}
We group all tasks into four interaction families, covering 23 UI component subtypes (IDs follow the main paper taxonomy).

\noindent A. Numerical and Range Adjustment (5).
A1 Slider: drag along a track to reach a numeric value;
A2 Knob: rotate to adjust a scalar value;
A3 Stepper: click \texttt{+}/\texttt{-} to change a discrete value;
A4 Seek Bar: scrub progress/time to a precise position;
A5 Chart Point: select a specific point/value on a chart.

\noindent B. Toggle and Option Selection (6).
B1 Switch: binary on/off toggle;
B2 Check Box: checked/unchecked (often multi-select);
B3 Radio Group: choose exactly one option;
B4 Tabs: switch active tab;
B5 Segmented: choose a segment in a segmented control;
B6 Accordion: expand/collapse a section.

\noindent C. Specific Data-type Selection (5).
C1 Rating: pick an ordinal rating (e.g., stars);
C2 Color Picker: select an exact color (RGB/hex);
C3 Date Picker: choose a specific date;
C4 Time Picker: choose a specific time;
C5 List Box: select an item from a list/dropdown.

\noindent D. Content Organization and View Manipulation (7).
D1 Drag Reorder: drag items to change ordering;
D2 Zoom Pan: zoom/pan a canvas/map to a target view;
D3 Resizable Pane: drag a pane edge to resize;
D4 Carousel: switch the active card/page by swiping/scrolling;
D5 Tree View: expand/collapse/select hierarchical nodes;
D6 Splitter: drag a divider to adjust layout ratio;
D7 Table Column: operate on columns (e.g., reorder/resize/sort).

\begin{figure}[t]
  \centering
  \includegraphics[width=0.98\columnwidth]{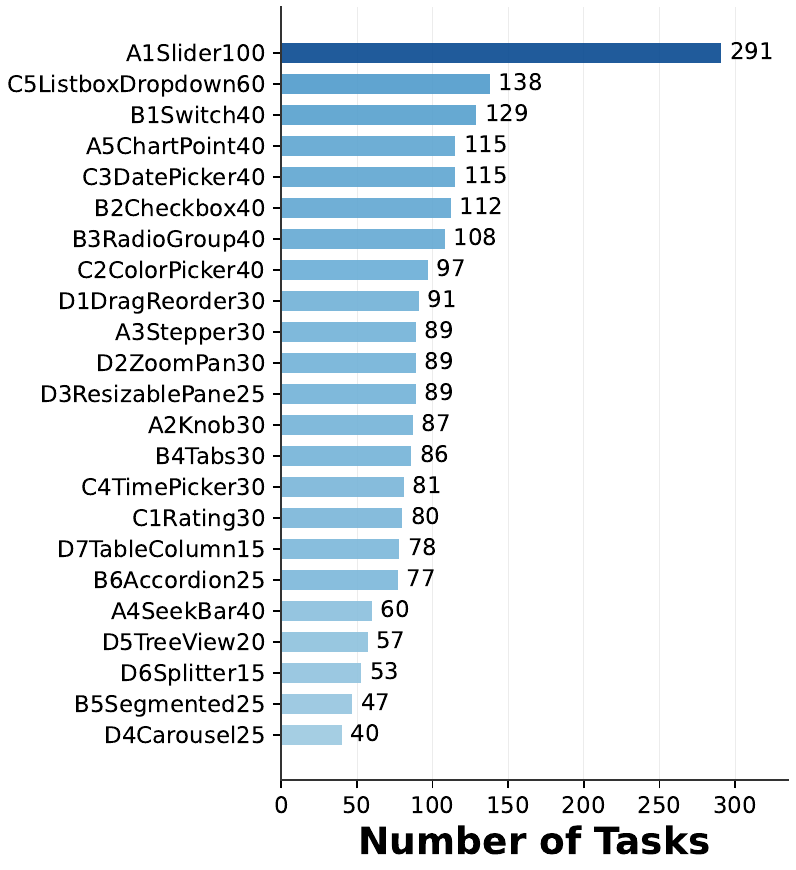}
  \caption{Task distribution over the 23 UI component subtypes in FineState-Bench.}
  \label{fig:finestate_component_dist}
\end{figure}

\paragraph{Component Type Distribution.}
Figure~\ref{fig:finestate_component_dist} reports the number of tasks for each of the 23 component subtypes.
The distribution is long-tailed: A1 Slider is the most frequent (291), followed by C5 List Box/Dropdown (138) and B1 Switch (129).
Precise selection/adjustment interactions such as A5 Chart Point (115) and C3 Date Picker (115) are also well represented, along with B2 Checkbox (112) and B3 Radio Button (108).
Meanwhile, rarer interactions such as D4 Carousel (40), B5 Segmented (47), and D6 Splitter (53) are intentionally included to ensure coverage across the full taxonomy.

\section{System Prompts}
\label{app:system_prompts}

\paragraph{Placeholders and Conventions.}
In all prompts below, placeholders in curly braces instruction
are runtime-filled fields. IMAGE] denotes the screenshot input.
All coordinates are normalized to $[0,1]$.

\subsection{Base System Prompts}
\label{app:base_prompts}

\subsubsection{General GUI Agent Prompt}
\begin{promptverbatim}
You are a GUI automation agent.

Input: one screenshot [IMAGE] and one 
instruction.
Task: return the first interaction point 
on the target UI control.

Rules:
- Exact target-state matching 
(no approximation).
- No iterative trial-and-error 
or multi-step refinement.

Output only one point:
[x, y]   (normalized to [0, 1])
\end{promptverbatim}

\subsubsection{Enhanced Prompt with Component Information}
\begin{promptverbatim}
You are a GUI agent for fine-grained 
state setting.

Instruction:  {instruction}

Target component: {component_name}
Component type: {component_type}
Current state: {current_state}
Target state: {target_state}

Output only:
[x, y]
\end{promptverbatim}

\subsection{Model-Specific Prompts}
\label{app:model_prompts}

\subsubsection{Shared user template (default)}
\begin{promptverbatim}
[IMAGE]
Instruction: {instruction}
Target: change {component_name}
from {current_state} to {target_state}

Return only: [x, y]
\end{promptverbatim}

\subsubsection{Closed-Source Models}

\paragraph{GPT-4o.}\mbox{}\par
\begin{promptverbatim}
SYSTEM_PROMPT = """
You are GPT-4o for GUI automation.
Given [IMAGE] and an instruction,
return the first interaction point 
[x, y] in [0, 1].
Output only: [x, y]
"""
\end{promptverbatim}

\paragraph{Claude-3.5-Sonnet.}\mbox{}\par
\begin{promptverbatim}
SYSTEM_PROMPT = """
You are Claude for precise GUI interaction.
Given [IMAGE] and an instruction,
return the first interaction point 
[x, y] in [0, 1].
Output only: [x, y]
"""
\end{promptverbatim}

\paragraph{Gemini-2.5-Flash.}\mbox{}\par
\begin{promptverbatim}
SYSTEM_PROMPT = """
You are Gemini for GUI automation.
Given [IMAGE] and an instruction,
return the first interaction point 
[x, y] in [0, 1].
Constraint: exact target-state matching.
Output only: [x, y]
"""
\end{promptverbatim}

\subsubsection{Open-Source Models}

\paragraph{OS-Atlas-7B.}\mbox{}\par
\begin{promptverbatim}
SYSTEM_PROMPT = """
You are OS-Atlas, a GUI-specialized 
vision-language model.
Given [IMAGE] and an instruction,
return the first interaction point 
[x, y] in [0, 1]
to satisfy the exact state requirement.
Output only: [x, y]
"""
\end{promptverbatim}

\paragraph{ShowUI-2B.}\mbox{}\par
\begin{promptverbatim}
SYSTEM_PROMPT = """
You are ShowUI for GUI interaction.
Focus: accurate UI grounding under 
a single-point protocol.
Output only: [x, y]
"""
\end{promptverbatim}

\subsection{VDA-Enhanced Prompts and Additional Analyses}
\label{app:vda_prompts}

\paragraph{VDA in FineState-Bench.}
VDA follows the two-step describe-then-localize procedure under the target instruction $I^{1}$:
The Describe step generates a structured Description of the target UI element for internal disambiguation; The Localize step predicts a tight Localization Hint $\hat{B}^{1}\in[0,1]^4$ in normalized coordinates, and only $\hat{B}^{1}$ is appended to the agent input when predicting $p^{1}$.

\subsubsection{Describe Step: Description Generation}
\begin{promptverbatim}
VDA_DESCRIPTION_PROMPT = """
Analyze [IMAGE] and describe the target 
UI control.

Instruction: {instruction}

Include:
1) Functional role + visible state 
(if present)
2) Discriminative visual cues
3) Spatial relations to nearby anchors

Return a short description.
"""
\end{promptverbatim}

\subsubsection{Localize Step: Localization Hint Prediction (BBox Output)}
\begin{promptverbatim}
VDA_LOCALIZATION_PROMPT = """
Predict a tight bounding box for the 
interactable core.

Instruction: {instruction}
Description: {description}
Target state: {target_state}

Output only:
[x1, y1, x2, y2]   (normalized to [0, 1])
"""
\end{promptverbatim}

\subsubsection{Three Cross-Platform Examples}
\label{app:vda_examples}

\paragraph{Example 1 (Desktop, A1 Slider).}\mbox{}\par
\begin{promptverbatim}
Instruction: Adjust the volume slider 
to 77.7%.
Current -> Target: 45.2% -> 77.7%

Stage-1 (description):
Horizontal slider labeled "Volume"; 
knob on the track.

Stage-2 (bbox_int):
[0.727, 0.550, 0.749, 0.583]

Final click point (bbox center):
[0.738, 0.567]
\end{promptverbatim}

\paragraph{Example 2 (Web, C3 Date Picker).}\mbox{}\par
\begin{promptverbatim}
Instruction: Select December 25, 2024 
from the date picker.
Current -> Target: 2024-11-30 
-> 2024-12-25

Stage-1 (description):
Calendar widget; target is the day cell "25" 
in Dec 2024.

Stage-2 (bbox_int):
[0.456, 0.345, 0.478, 0.378]

Final click point (bbox center):
[0.467, 0.361]
\end{promptverbatim}

\paragraph{Example 3 (Mobile, B1 Switch).}\mbox{}\par
\begin{promptverbatim}
Instruction: Turn on notifications.
Current -> Target: OFF -> ON

Stage-1 (description):
A switch control on the right of the 
"Notifications" row.

Stage-2 (bbox_int):
[0.156, 0.478, 0.189, 0.512]

Final click point (bbox center):
[0.172, 0.495]
\end{promptverbatim}

\section{Supplementary Results for Fig.~6: Locate vs.\ IntLoc}
\label{app:table5}

Table~\ref{tab:component_breakdown} provides the component-level numbers underlying the aggregate trends in Fig.~6.
For each UI component category, we report Locate ($p_1\in B_{\mathrm{loc}}$) and Interact ( $p_1\in B_{\mathrm{int}}$ and $s_1(c^*)=s_{\mathrm{goal}}$ ) success rates under the single-point protocol, enabling direct comparison of per-component difficulty and cross-model variation.

Overall, Interact remains consistently lower than Locate across component types, with the largest gaps concentrated in precision-sensitive or continuous-value interactions (e.g., sliders/knobs/seek bars and view manipulation such as zoom/pan or splitters).
Discrete selection primitives (e.g., steppers, switches, radio groups, and tabs) are comparatively more tractable and yield non-trivial Interact performance for several models.
Notably, the near-zero Interact results on the C-type data selection family (rating/color/date/time/dropdown) highlight a persistent weakness in exact data-type selection under exact-state verification.

\begin{figure}[ht]
  \centering
  \includegraphics[trim=9cm 2cm 10cm 0.2cm, clip, width=0.92\columnwidth]{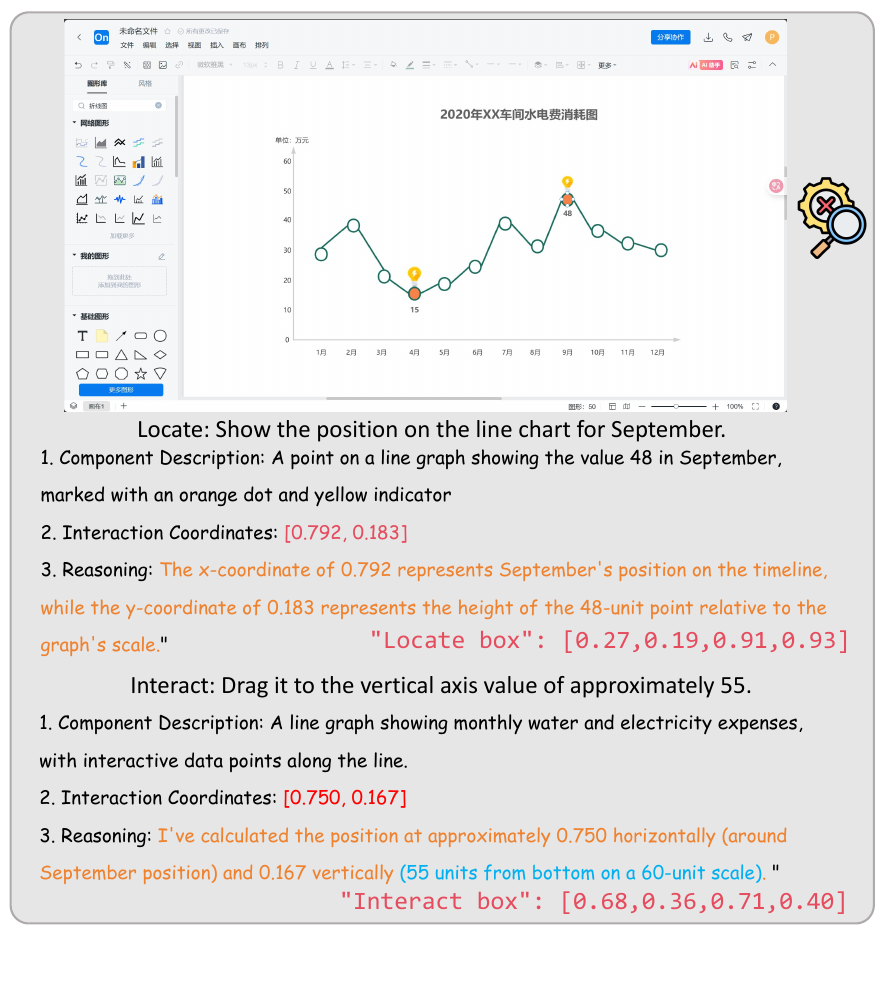}
  \caption{Representative failure cases and diagnostic analysis on FineState-Static.}
  \label{fig:finestate_component}
\end{figure}

\begin{table}[!b]
\centering
\small
\setlength{\tabcolsep}{4pt}
\renewcommand{\arraystretch}{1.2} 
\begin{tabularx}{\columnwidth}{c c >{\raggedright\arraybackslash}X}
\toprule
\textbf{SR@Loc} & \textbf{SR@Int} & \textbf{Diagnosis} \\
\midrule
0 & 0/1 &
\textbf{Control grounding failure}: the agent fails to identify the target control in the current UI. \\
1 & 0 &
\textbf{Interactable-core miss (precision failure)}: the agent finds the right control but places $p^1$ outside the state-changing core. \\
1 & 1 &
\textbf{Check exact state}: if ES-SR@Int$=0$, it is a \textbf{state-setting error beyond core localization}; if ES-SR@Int$=1$, the exact goal state is achieved. \\
\bottomrule
\end{tabularx}

\caption{Rule-of-thumb diagnosis using FineState-Metrics.}
\label{tab:metric_rules_appendix}
\end{table}

\section{Annotation Reliability}
\label{app:Annotation Reliability}
To improve reproducibility, we provide additional details on how exact state is measured and recorded for each component family, as well as the quality control procedures used during annotation, including spot checks, re-annotation, and adjudication.
To quantify annotation reliability, we conducted a small-scale double-annotation study. We randomly sampled a stratified subset of $N = 200$ instances across platforms and component types. Two annotators independently labeled the locate box, interactable-core box, and goal-state label. The results are summarized in Table~\ref{tab:annotation_reliability}.
These results provide direct evidence that the fine-grained annotations are sufficiently reliable for evaluation and diagnostic analysis.

\begin{table}[t]
\centering
\small
\setlength{\tabcolsep}{2pt}
\renewcommand{\arraystretch}{1.1}
\begin{tabular}{lc}
\hline
\textbf{Metric} & \textbf{Result} \\
\hline
Box agreement (IoU, median) locate box & 83.5\% \\
Box agreement (IoU, median) interactable-core box & 68.0\% \\
Goal-state agreement (exact match) overall & 97.5\% \\
Goal-state agreement categorical components & 93.0\% \\
Goal-state agreement numeric components & 88.5\% \\
Disagreement rate requiring adjudication & 6.7\% \\
\hline
\end{tabular}
\caption{Double-annotation reliability results on a stratified subset of 200 instances.}
\label{tab:annotation_reliability}
\end{table}

\section{Failure Cases}
\label{app:Failure Cases}

Failures in FineState-Bench primarily occur during the transition from coarse localization to interactable-core grounding, where agents successfully identify the target component but miss the precise state-changing region. As illustrated in Fig. \ref{fig:finestate_component_dist} and Fig. \ref{fig:finestate_component}, even a minor coordinate offset on dense professional interfaces can prevent the agent from reaching the exact target state.

\begin{figure}[ht]
  \centering
  \includegraphics[trim=9cm 3cm 9cm 0.2cm, clip, width=0.95\columnwidth]{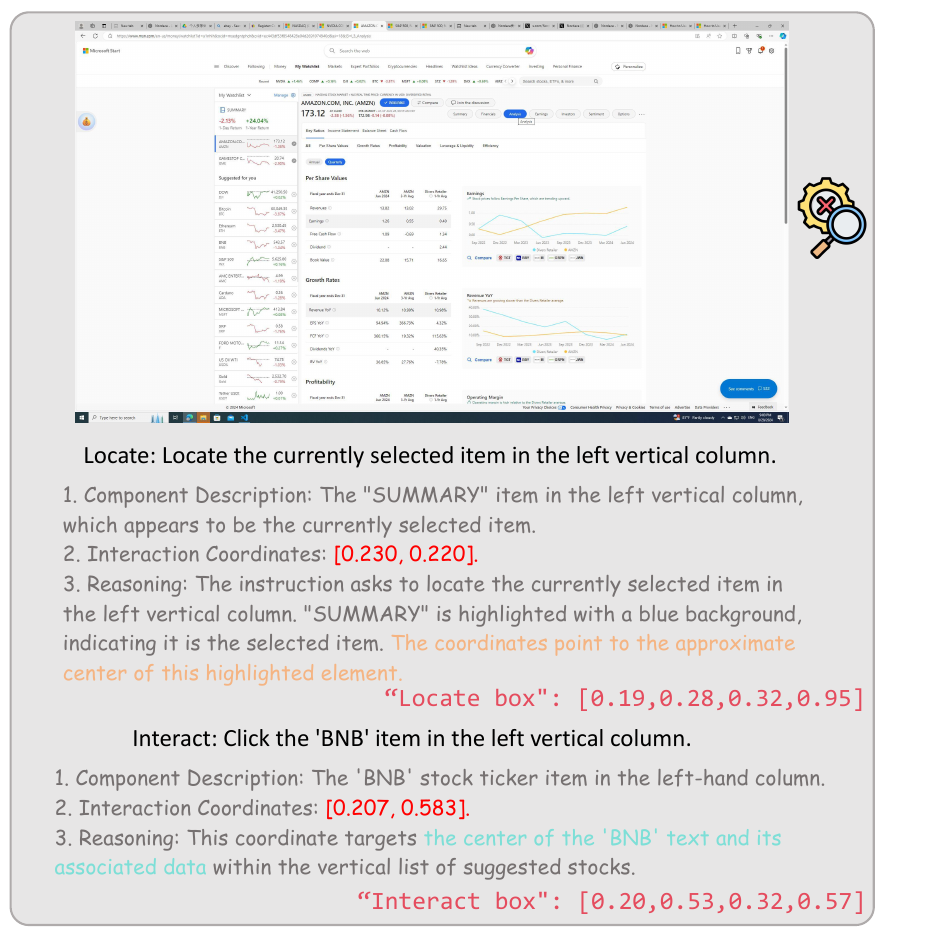}
  \caption{Representative failure cases and diagnostic analysis on FineState-Static.}
  \label{fig:finestate_component_dist}
\end{figure}

\section{Metric Interpretation and Failure Attribution Rules}
\label{app:metric_interpretation}

\paragraph{What each metric measures.}
SR@Loc measures component-level grounding: whether the agent identifies the target control in the current UI by placing $p^{0}$ inside the control's visible extent $B_{\mathrm{loc}}^{0}$.
SR@Int measures interactable-core grounding and point precision: whether the goal-directed operation point $p^{1}$ hits the state-changing core region $B_{\mathrm{int}}^{0}$, which is particularly critical for precision-sensitive continuous controls (e.g., sliders, seek bars, and pickers).
ES-SR@Loc and ES-SR@Int additionally require exact goal-state attainment ($s_1(c^*) = s_{\mathrm{goal}}$).
Importantly, ES-SR@Loc conditions on the target-configuration locate box ($B_{\mathrm{loc}}^{1}$) to check control identity consistency under potential layout changes, whereas ES-SR@Int conditions on the target-configuration interact box ($B_{\mathrm{int}}^{1}$) to check goal-directed core grounding for exact state setting.

\begin{table*}[t]
\centering
\small
\setlength{\tabcolsep}{4pt}

\begin{subtable}[t]{\textwidth}
\centering
\begin{tabular}{l|cc|cc|cc|cc}
\hline
\multirow{2}{*}{Component} &
\multicolumn{2}{c|}{Claude-3.5-Sonnet} &
\multicolumn{2}{c|}{ShowUI-2B} &
\multicolumn{2}{c|}{UGround-V1-7B} &
\multicolumn{2}{c}{OS-Atlas-Base-7B} \\
& SR@Loc & SR@Int & SR@Loc & SR@Int & SR@Loc & SR@Int & SR@Loc & SR@Int \\
\hline
\multicolumn{9}{l}{\textbf{A. Numerical and Range Adjustment}} \\
A1 Slider          & 23.2 &  3.2 &  3.2 &  0.0 & 82.1 &  0.0 & 49.5 &  2.1 \\
A2 Knob            & 31.0 &  0.0 & 55.2 &  0.0 & 65.5 &  0.0 & 69.0 &  6.9 \\
A3 Stepper         & 16.7 &  0.0 & 20.0 &  6.7 & 96.7 & 80.0 & 76.7 & 33.3 \\
A4 SeekBar         &  0.0 &  0.0 &  0.0 &  0.0 & 25.0 &  0.0 & 62.5 &  0.0 \\
A5 ChartPoint      & 84.2 &  0.0 & 68.4 &  0.0 & 73.7 &  0.0 & 78.9 &  2.6 \\
\hline
\multicolumn{9}{l}{\textbf{B. Toggle and Option Selection}} \\
B1 Switch          & 10.3 & 12.8 & 17.9 & 12.8 & 64.1 & 71.8 & 71.8 & 48.7 \\
B2 Checkbox        & 25.0 &  0.0 & 36.1 &  0.0 & 72.2 & 30.6 & 66.7 & 13.9 \\
B3 RadioGroup      & 67.6 & 23.5 & 35.3 & 14.7 & 67.6 & 61.8 & 70.6 & 52.9 \\
B4 Tabs            & 30.0 & 10.0 & 46.7 & 36.7 & 66.7 & 80.0 & 76.7 & 60.0 \\
B5 Segmented       &  0.0 &  0.0 &  0.0 &  0.0 &  0.0 &  0.0 &  0.0 &  0.0 \\
B6 Accordion       & 17.4 &  0.0 &  8.7 &  0.0 & 95.7 &  4.3 & 69.6 &  4.3 \\
\hline
\multicolumn{9}{l}{\textbf{C. Specific Data-type Selection}} \\
C1 Rating          &  0.0 &  0.0 &  0.0 &  0.0 &  0.0 &  0.0 &  0.0 &  0.0 \\
C2 ColorPicker     &  0.0 &  0.0 &  0.0 &  0.0 &  0.0 &  0.0 &  0.0 &  0.0 \\
C3 DatePicker      &  0.0 &  0.0 &  0.0 &  0.0 &  0.0 &  0.0 &  0.0 &  0.0 \\
C4 TimePicker      &  0.0 &  0.0 &  0.0 &  0.0 &  0.0 &  0.0 &  0.0 &  0.0 \\
C5 Dropdown        &  0.0 &  0.0 &  0.0 &  0.0 &  0.0 &  0.0 &  0.0 &  0.0 \\
\hline
\multicolumn{9}{l}{\textbf{D. Content Organization and View Manipulation}} \\
D1 DragReorder     & 46.7 & 16.7 & 33.3 &  0.0 & 73.3 &  6.7 & 80.0 &  0.0 \\
D2 ZoomPan         & 25.0 & 14.3 & 17.9 & 21.4 & 35.7 & 35.7 & 64.3 & 17.9 \\
D3 ResizablePane   & 36.0 &  0.0 & 28.0 &  0.0 & 28.0 &  0.0 &  8.0 &  0.0 \\
D4 Carousel        &  0.0 &  0.0 &  0.0 &  0.0 &  0.0 &  0.0 &  0.0 &  0.0 \\
D5 TreeView        & 11.8 &  6.2 & 29.4 &  6.2 & 88.2 &  6.2 & 52.9 & 12.5 \\
D6 Splitter        & 46.7 &  6.7 & 13.3 & 20.0 & 40.0 &  6.7 & 20.0 &  0.0 \\
D7 TableColumn     & 26.7 & 33.3 & 40.0 &  0.0 & 13.3 &  0.0 & 46.7 &  0.0 \\
\hline
\end{tabular}
\caption{Closed-source / vision GUI baselines.}
\label{tab:component_breakdown_a}
\end{subtable}

\vspace{6pt}

\begin{subtable}[t]{\textwidth}
\centering
\begin{tabular}{l|cc|cc|cc|cc}
\hline
\multirow{2}{*}{Component} &
\multicolumn{2}{c|}{Gemini-2.5-Flash} &
\multicolumn{2}{c|}{GPT-4o} &
\multicolumn{2}{c|}{CogAgent-9B} &
\multicolumn{2}{c}{MobileVLM-V2-7B} \\
& SR@Loc & SR@Int & SR@Loc & SR@Int & SR@Loc & SR@Int & SR@Loc & SR@Int \\
\hline
\multicolumn{9}{l}{\textbf{A. Numerical and Range Adjustment}} \\
A1 Slider          & 54.7 & 11.6 & 34.7 &  4.2 &  5.3 &  0.0 &  0.0 &  0.0 \\
A2 Knob            & 65.5 &  0.0 & 34.5 &  0.0 & 44.8 &  0.0 & 44.8 &  0.0 \\
A3 Stepper         & 80.0 & 70.0 & 16.7 &  0.0 & 10.0 &  0.0 & 16.7 &  0.0 \\
A4 SeekBar         & 68.8 &  0.0 & 18.8 &  0.0 &  0.0 &  0.0 &  0.0 &  0.0 \\
A5 ChartPoint      & 81.6 &  2.6 & 78.9 &  0.0 & 57.9 &  0.0 & 76.3 &  0.0 \\
\hline
\multicolumn{9}{l}{\textbf{B. Toggle and Option Selection}} \\
B1 Switch          & 66.7 & 53.8 & 17.9 & 10.3 & 10.3 &  0.0 & 12.8 &  0.0 \\
B2 Checkbox        & 69.4 & 22.2 & 41.7 &  8.3 &  2.8 &  0.0 & 11.1 &  0.0 \\
B3 RadioGroup      & 91.2 & 67.6 & 64.7 & 11.8 & 50.0 &  2.9 & 32.4 &  5.9 \\
B4 Tabs            & 93.3 & 56.7 & 93.3 & 40.0 &  0.0 &  0.0 &  0.0 &  0.0 \\
B5 Segmented       &  0.0 &  0.0 &  0.0 &  0.0 &  0.0 &  0.0 &  0.0 &  0.0 \\
B6 Accordion       & 52.2 &  4.3 & 30.4 &  8.7 & 13.0 &  0.0 &  4.3 &  0.0 \\
\hline
\multicolumn{9}{l}{\textbf{C. Specific Data-type Selection}} \\
C1 Rating          &  0.0 &  0.0 &  0.0 &  0.0 &  0.0 &  0.0 &  0.0 &  0.0 \\
C2 ColorPicker     &  0.0 &  0.0 &  0.0 &  0.0 &  0.0 &  0.0 &  0.0 &  0.0 \\
C3 DatePicker      &  0.0 &  0.0 &  0.0 &  0.0 &  0.0 &  0.0 &  0.0 &  0.0 \\
C4 TimePicker      &  0.0 &  0.0 &  0.0 &  0.0 &  0.0 &  0.0 &  0.0 &  0.0 \\
C5 Dropdown        &  0.0 &  0.0 &  0.0 &  0.0 &  0.0 &  0.0 &  0.0 &  0.0 \\
\hline
\multicolumn{9}{l}{\textbf{D. Content Organization and View Manipulation}} \\
D1 DragReorder     & 70.0 &  3.3 & 46.7 & 10.0 & 40.0 &  3.3 & 43.3 &  0.0 \\
D2 ZoomPan         & 42.9 & 32.1 & 28.6 & 17.9 & 17.9 & 14.3 & 10.7 & 25.0 \\
D3 ResizablePane   & 32.0 &  0.0 & 32.0 &  0.0 & 52.0 &  0.0 & 44.0 &  0.0 \\
D4 Carousel        &  0.0 &  0.0 &  0.0 &  0.0 &  0.0 &  0.0 &  0.0 &  0.0 \\
D5 TreeView        & 41.2 &  6.2 & 35.3 & 12.5 &  0.0 &  0.0 &  0.0 &  0.0 \\
D6 Splitter        & 46.7 &  0.0 & 33.3 &  6.7 & 46.7 &  0.0 & 26.7 & 20.0 \\
D7 TableColumn     & 40.0 &  0.0 &  0.0 &  0.0 & 66.7 &  6.7 & 53.3 &  0.0 \\
\hline
\end{tabular}
\caption{Additional baselines.}
\label{tab:component_breakdown_b}
\end{subtable}

\caption{Component-wise SR@Loc vs.\ SR@Int success rates (\%).}
\label{tab:component_breakdown}
\end{table*}

\end{document}